\documentclass[letterpaper]{article}
\usepackage[preprint]{aaai2027}
\usepackage[hyphens]{url}
\usepackage{graphicx}
\usepackage{natbib}
\usepackage{caption}
\usepackage{amsmath,amssymb,bm}
\usepackage{booktabs}
\usepackage{eso-pic}

\newcommand{\JPMDisclaimer}{%
  \par\smallskip
  \begingroup
  \footnotesize
  \noindent\textbf{Disclaimer:} This paper was prepared for informational purposes by the LLM Suite group of JP Morgan Chase and its affiliates (\textquoteleft{}JPMC\textquoteright{}) and is not a product of the Research Department of JP Morgan. JP Morgan makes no representation, warranty or undertaking whatsoever and disclaims all liability for the completeness, accuracy or reliability of the information contained herein. This document is not intended as investment research or investment advice, or a recommendation, offer or solicitation for the purchase or sale of any security, financial instrument, financial product or service, or to be used in any way for evaluating the merits of participating in any transaction, and shall not constitute a solicitation under any jurisdiction to any person, if such solicitation under such jurisdiction or to such person would be unlawful.
  \par
  \endgroup
  \medskip
}

\newcommand{\JPMCopyrightNotice}{%
  \copyright{} 2026 JP Morgan Chase \& Co. All rights reserved%
}
\AddToShipoutPictureFG{%
  \AtPageLowerLeft{%
    \put(0,22){%
      \makebox[\paperwidth][c]{\normalfont\footnotesize\JPMCopyrightNotice}%
    }%
  }%
}

\title{Do Frontier Model Ingredients Add Up?\\A Replicated Factorial Study of Small Language Models}
\author{Biswa Sengupta}
\affiliations{
LLM Suite Team, JPMorgan Chase \& Co.\\
biswa.sengupta@jpmorgan.com
}
\begin{document}
\maketitle
\begin{abstract}
Building a small language model requires choosing which published improvements to combine. We compare five changes in the same model, using the same data and training budget. We test selected combinations and repeat each one with a second random seed. Mixture-of-experts layers, Muon optimisation and a Gated DeltaNet hybrid give the largest gains. Extra normalisation also helps, while an attention gate with a matched parameter budget adds little on average. The gains mostly combine, but small interactions between changes appear again in the repeated runs. Repetition also reveals benefits that an initial statistical estimate treated as noise. These findings apply to our small-model training setting; the expert models also have greater total capacity. The practical lesson is to test promising components together and repeat the comparisons before deciding which ones to keep.
\end{abstract}
\JPMDisclaimer

\section{Introduction}

Published improvements to language models are usually measured against different baselines. A builder must decide whether they will still help when used together. Two changes may provide the same benefit, or one may change how well the other works. We call the latter an \emph{interaction}.

We compare five changes in one implementation. G adds a sigmoid attention-output gate \citep{qiu2025gated}; M uses Muon to update hidden weight matrices \citep{jordan2024muon,liu2025muon}; S normalises both the input and output of each sub-layer \citep{ding2021cogview,kim2025periln}; E adds small routed experts \citep{dai2024deepseekmoe,wang2024auxfree,deepseek2024v3}; and H replaces six attention layers with Gated DeltaNet \citep{yang2024deltanet,yang2025gdn}.

Testing every combination requires 32 configurations before any repeats. We use a classical fractional factorial design: a balanced selection of 16 combinations that allows several comparisons to share the same runs \citep{fisher1935design,box1961fractional}. We then repeat each configuration with a second seed. A \emph{contrast} is a difference between two group averages; for a single switch, it compares average loss with the switch on and off.

The largest gains mostly combine, while smaller interactions also repeat. The design has an important limit: some effects are indistinguishable in the selected comparisons. This is called \emph{aliasing}. Repeating the runs measures variability but does not remove that ambiguity. We report these limits alongside the results so that the study can guide component selection at the tested scale.

\section{Related Work}

\citet{geiping2023cramming} study how to train masked language models within a single-GPU-day budget. \citet{dodge2019show} show why comparisons should account for the compute spent developing a model. \citet{wen2025fantastic} show how optimiser rankings depend on tuning and the final training budget. Our study compares several choices together and repeats every selected combination.

Factorial language-model studies also predate ours. ATMA selects a long-context recipe with a complete 120-cell sweep of attention, memory, and training choices \citep{akbar2026atma}. Our contribution is a repeated comparison of five different types of change in a small decoder. Both the components and the experimental-design method are established.

\section{Model, Switches, and Training}

\subsection{Common protocol}

The baseline is an eight-layer decoder of width 512 with eight 64-dimensional heads, SwiGLU feed-forward width 1,408, pre-RMSNorm, query/key normalisation, rotary positions, and tied embeddings. Its vocabulary has 50,304 entries and it contains 51,455,488 parameters. These are standard Transformer choices \citep{vaswani2017attention,shazeer2020glu,zhang2019rmsnorm,henry2020qknorm,su2024roformer}.

We train on FineWeb \citep{penedo2024fineweb}, using GPT-2 byte-pair tokenisation and the NanoGPT-speedrun shard format \citep{radford2019gpt2,jordan2024speedrun}. Sequences contain 1,024 tokens. Four micro-batches of 16 sequences give 65,536 tokens per step; 4,577 steps reach the nominal 300M-token budget. Final validation measures next-token cross-entropy over 10,485,760 held-out tokens. Seeds change initial weights and the contiguous range of training shards. Configurations with the same seed receive identical batches, checked by first-batch hashes. Loss is in nats per token; lower is better.

The learning rate rises for 300 steps, remains constant, then falls linearly over the last 20\% of training. AdamW \citep{loshchilov2019adamw} uses momentum coefficients $(0.9,0.95)$ and gradient clipping at 1.0. Preliminary sweeps select one peak rate per optimiser family: $0.002$ for AdamW and $0.016$ for Muon. For the relevant matrix groups, decay preserves the baseline product $\eta\lambda=2\times10^{-4}$ of peak rate and weight decay. Muon updates hidden matrices with five Newton--Schulz iterations, momentum 0.95, and the Moonlight update scale. Embeddings, norms, routers, gates, and convolution kernels retain AdamW. Thus M compares complete optimiser recipes, including rates and parameter groups. We do not tune each architecture separately.

\begin{table}[t]
\centering
\begin{tabular}{lp{0.75\columnwidth}}
\toprule
Switch & Change from the baseline \\
\midrule
G & Attention-output sigmoid gate; narrower feed-forward layers to keep active size similar. \\
M & Muon for hidden matrices; AdamW for other groups, with the chosen family settings. \\
S & Embedding RMSNorm and output RMSNorm before each sub-layer joins the residual stream. \\
E & Layers 2--8: 32 routed experts, top-3 selection, one shared expert; width 352 each. \\
H & Gated DeltaNet in layers 1--3 and 5--7; global attention in layers 4 and 8. \\
\bottomrule
\end{tabular}
\caption{The five switches, each of which can be on or off. They share one training implementation. Active parameter counts are approximately matched; total counts are not.}
\label{tab:switches}
\end{table}

\subsection{What each switch changes}

Table~\ref{tab:switches} defines the switches. G adds a $512\times512$ projection to each attention layer. We pay for it by reducing feed-forward width from 1,408 to 1,240 and active expert width from 352 to 309. G therefore tests whether the gate is a useful way to spend the parameter budget.

For E, a sigmoid score and a balancing bias select the top three routed experts. The bias receives no gradient. Output weights use the original scores, normalised over the selected experts and multiplied by $\sqrt{3}$; the shared expert is always active. A sign update increases the selection bias of underused experts. Its step size is $10^{-2}$ for 1,000 steps and $10^{-3}$ thereafter. A sequence-wise balance term has coefficient $10^{-4}$. The model has about 161.3M total and 51.6M active parameters. Its gain can therefore reflect both routing and roughly three times the total capacity.

For H, unit-normalised keys drive the gated delta update,
\begin{equation}
S_t=\alpha_t(I-\beta_t k_t k_t^\top)S_{t-1}
       +\beta_t k_t v_t^\top,
\label{eq:gdn-factorial}
\end{equation}
where $\alpha_t,\beta_t\in(0,1)$ and each head stores a $64\times64$ state \citep{yang2025gdn}. Causal width-four convolutions filter the projected queries, keys and values before the memory update. The two remaining attention layers use no positional encoding (NoPE) \citep{kazemnejad2023nope}. H tests this whole hybrid recipe.

\section{Experimental Design and Analysis}

\subsection{Choosing the combinations}

Code each switch as $x_f\in\{-1,+1\}$. We run the 16 configurations satisfying
\begin{equation}
x_Gx_Mx_Sx_Ex_H=+1,\qquad I=GMSEH.
\label{eq:fraction}
\end{equation}
These comprise five single-switch configurations, ten three-switch configurations, and the all-five configuration. The all-off baseline is outside this selection. We use it only for comparisons labelled as standalone. Two seeds per configuration give 32 core training runs, or about 9.6B training tokens. This count excludes preliminary sweeps, diagnostics, and the extra hybrid repeats used below.

Let $y_{cr}$ be final validation loss for configuration $c$ and seed $r$. We describe it using a model with single-switch and pairwise terms:
\begin{equation}
y_{cr}=\beta_0+\sum_f\beta_f x_{fc}
 +\sum_{f<g}\beta_{fg}x_{fc}x_{gc}+\varepsilon_{cr}.
\label{eq:model}
\end{equation}
There are 16 coefficients and 16 configurations. The design balances the comparison columns so that they are orthogonal. For a single-switch or pair-product column $z_{Ac}$, the contrast and its two-seed mean are
\begin{equation}
\widehat e_A^{(r)}=\frac{2}{16}\sum_{c=1}^{16}z_{Ac}y_{cr},
\qquad
\overline e_A=\frac{\widehat e_A^{(0)}+\widehat e_A^{(1)}}{2}.
\label{eq:effect}
\end{equation}
Negative single-switch contrasts favour turning that switch on.

This is a resolution-V design \citep{box1961fractional}: single-switch and pairwise effects can be separated from each other, but some higher-order effects remain mixed in. For example, $G\equiv MSEH$ and $GM\equiv SEH$. The measured G contrast includes the four-switch effect MSEH; the measured GM contrast includes the three-switch effect SEH. Interpreting them as G or GM alone requires those higher-order effects to be small. Replication cannot separate them. A small measured contrast could also hide two large contributions with opposite signs.

\subsection{Measuring variation between runs}

With one seed, Equation~\ref{eq:model} fits as many coefficients as there are observations. It leaves no residual estimate of run noise. We initially use Lenth's pseudo-standard error (PSE), which estimates noise from the spread of effects under the assumption that most are inactive \citep{lenth1989quick}. It gives PSE $0.0102$ nats. This estimate comes from the effects themselves, rather than repeated runs.

For the repeated-run estimate, let $s_H^2$ be the sample variance of five completed H-only runs. Let $\delta_c=y_{c1}-y_{c0}$ for the other 15 configurations. Combining these gives
\begin{equation}
\widehat\sigma^2=
\frac{4s_H^2+\tfrac12\sum_{c\ne H}\delta_c^2}{19},
\qquad \widehat\sigma=0.0036.
\label{eq:noise}
\end{equation}
H's first pair is counted only once. If run errors are independent and have equal variance, the standard error of a two-seed mean contrast is
\begin{equation}
\operatorname{SE}(\overline e_A)=\frac{\widehat\sigma}{2\sqrt{2}}
 \simeq0.0013.
\label{eq:se}
\end{equation}
With 19 nominal degrees of freedom and a \v{S}id\'ak adjustment for 15 comparisons, this gives a simultaneous 95\% reference margin of $\pm0.0043$ nats. This is a guide under the stated assumptions. Errors may move together because configurations share data streams, and two seeds cannot estimate their full covariance or establish equal variances. We report both seeds so readers can also judge agreement directly.

\section{Results}

\subsection{The main gains repeat}

\begin{figure}[t]
\centering
\includegraphics[width=\columnwidth]{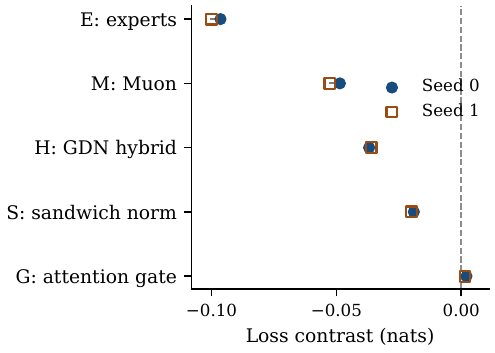}
\caption{Average loss differences for each switch at both seeds. Negative values favour the switch. Markers show estimates from different seeds, not uncertainty ranges. Each comparison can also include the corresponding four-switch effect because of the selected design.}
\label{fig:main-effects}
\end{figure}

Experts, Muon, and the hybrid give the largest average reductions in loss (Table~\ref{tab:contrasts}, Figure~\ref{fig:main-effects}). Extra normalisation also helps. Single-switch contrasts differ by at most 0.004 nats between seeds. The Pearson correlation is 0.996 across all 15 contrasts and 0.94 across interaction contrasts alone.

\begin{table}[t]
\centering
\begin{tabular}{lrrr}
\toprule
Contrast & Seed 0 & Seed 1 & Mean \\
\midrule
E & $-0.0964$ & $-0.1001$ & $-0.0983$ \\
M & $-0.0486$ & $-0.0526$ & $-0.0506$ \\
H & $-0.0369$ & $-0.0361$ & $-0.0365$ \\
S & $-0.0190$ & $-0.0198$ & $-0.0194$ \\
G & $+0.0021$ & $+0.0015$ & $+0.0018$ \\
\midrule
SE / GMH & $-0.0124$ & $-0.0099$ & $-0.0111$ \\
ME / GSH & $-0.0094$ & $-0.0090$ & $-0.0092$ \\
GE / MSH & $-0.0073$ & $-0.0063$ & $-0.0068$ \\
GS / MEH & $-0.0063$ & $-0.0041$ & $-0.0052$ \\
MS / GEH & $+0.0014$ & $-0.0021$ & $-0.0004$ \\
EH / GMS & $+0.0012$ & $-0.0005$ & $+0.0003$ \\
GH / MSE & $+0.0042$ & $-0.0018$ & $+0.0012$ \\
SH / GME & $+0.0027$ & $+0.0033$ & $+0.0030$ \\
GM / SEH & $+0.0108$ & $+0.0124$ & $+0.0116$ \\
MH / GSE & $+0.0106$ & $+0.0139$ & $+0.0122$ \\
\bottomrule
\end{tabular}
\caption{Loss contrasts in nats. A slash names effects that this design cannot separate. Single-switch rows also include four-switch effects. Means use unrounded losses, so they may differ from averages of the displayed values. The reference margin for a mean is $\pm0.0043$, under the noise assumptions in the text.}
\label{tab:contrasts}
\end{table}

The gate shows why the common comparison matters. Adding G alone to the all-off baseline reduces seed-0 loss by 0.019 nats. Across the selected combinations, its repeated mean is instead $+0.0018$, within the reference margin. Spending active parameters on this gate gives little average benefit here. This does not tell us whether an extra gate without a narrower feed-forward layer would help, or which other change reduces its benefit.

\subsection{Gains mostly combine, with small interactions}

The largest mean pairwise contrast is MH/GSE at $+0.0122$ nats. Adding the reference margin gives about 0.017 nats, below the experiment's practical bound of 0.02 nats. Even so, six pairwise means and four single-switch means exceed the reference margin. The three largest seed-0 interaction contrasts---SE/GMH, GM/SEH, and MH/GSE---keep their signs and exceed the single-seed reference margin of $\pm0.0060$ at seed 1.

The large gains therefore mostly combine, but the smaller departures matter when comparing modest refinements. If higher-order effects are negligible, a pairwise contrast of 0.012 means that turning one switch on changes the other's benefit by about 0.024 nats.

The all-five configuration has the lowest loss in the selected set at both seeds: 3.6792 and 3.6750 nats. The untested combinations or separately tuned recipes could do better. In particular, G's near-zero average does not show that removing it from the all-five model would help: that four-switch configuration was not run.

\subsection{Repeats reveal smaller effects}

Lenth's PSE implies a run standard deviation near 0.020 nats, using the single-seed contrast standard error $\sigma/2$. That is about six times the repeated-run estimate of 0.0036. The initial margin separates only E, M, and H from zero; the repeated-run reference also separates S and several interaction contrasts. Many effects in this experiment contain signal, so their spread overestimates noise. This explains the conservative initial result without implying that Lenth's method generally fails.

\section{Checking the Implementation}

Two numerical failures were found during development, repaired, and excluded from the final results. First, an RMSNorm GPU kernel let a position use information from the next token at a particular row-byte width. The resulting very low loss was invalid. Tests at the actual model width, changing future tokens at both odd and even positions, exposed the leak and checked the replacement formula. Smaller-width tests had missed it.

Second, the sigmoid router could produce finite outputs but non-finite gradients when selected scores were extremely small. Replaying a saved failing batch traced this to a squared denominator that underflowed during differentiation. We use the equivalent expression
\begin{equation}
g_i=\rho\,\operatorname{softmax}_{i\in\mathcal K}
              \big(\log\operatorname{sigmoid}(\ell_i)\big),
\label{eq:stable-router}
\end{equation}
with a stable log-sigmoid implementation and $\rho=\sqrt3$, instead of differentiating $\rho s_i/\sum_{j\in\mathcal K}s_j$ directly. The final comparison uses runs after both fixes.

\section{Limitations and Reproducibility}

\paragraph{Where the findings apply.} The study covers about 51.5M active parameters, one corpus, a 1,024-token context, and a nominal 300M-token budget. Training time, energy and inference speed are not matched. The results do not establish downstream or long-context gains, or behaviour at larger scales. Optimiser comparisons show why size, tuning and final budget can change rankings \citep{wen2025fantastic}.

\paragraph{What the comparisons measure.} E adds total capacity, G spends some feed-forward capacity on a gate, H changes both recurrent mixing and nearby design choices, and M includes several optimiser settings. These are comparisons of complete recipes. Also, the selected combinations cannot separate all effects. A complementary selection or targeted extra configurations would be needed to resolve the remaining ambiguity.

\paragraph{Variation and model selection.} Two seeds give little information about each configuration's full distribution. The noise pool assumes a common variance, and shared data streams may correlate errors. Both seeds are reported to make their agreement visible. The same validation corpus supports preliminary tuning and final comparisons, so these are development results rather than measurements on a fresh test set.

\paragraph{Reproduction.} An evidence ledger links definitions, settings, final losses and calculations to the source report and generated tables. It does not contain every raw log or checkpoint. The 32 core runs, learning-rate sweeps and extra H repeats must be counted separately. Reproduction should first check future-token causality at the model width, finite router gradients on extreme logits, and matching data streams. Equations~\ref{eq:effect}--\ref{eq:noise} then give the calculations from completed valid runs.

\section{Conclusion}

In this training setting, experts, Muon, a Gated DeltaNet hybrid and extra normalisation reduce average loss. A gate with a matched parameter budget adds little on average. The main gains mostly combine, but small interactions repeat. Comparing components together and repeating the runs gives a clearer basis for deciding which changes to keep.

\bibliography{references}
\end{document}